\documentclass[10pt,twocolumn,letterpaper]{article}

\usepackage{iccv}
\usepackage{times}
\usepackage{epsfig}
\usepackage{graphicx}
\usepackage{amsmath}
\usepackage{amssymb}
\usepackage{gensymb}

\usepackage[breaklinks=true,bookmarks=false]{hyperref}

\iccvfinalcopy 

\def\iccvPaperID{8} 

\ificcvfinal\fi
\begin{document}

\title{How to improve CNN-based 6-DoF camera pose estimation\\}

\author{Soroush Seifi \hspace{0,5cm} Tinne Tuytelaars\\
PSI, ESAT, KU Leuven\\
Kasteelpark Arenberg 10, 3001 Leuven, Belgium\\
{\tt\small \{sseifi, tinne.tuytelaars\}@esat.kuleuven.be}
}

\maketitle
\thispagestyle{empty}

\begin{abstract}
\vspace{-0,3cm}
Convolutional neural networks (CNNs) and transfer learning have recently been used for 6 degrees of freedom (6-DoF) camera pose estimation. 
While they do not reach the same accuracy as visual SLAM-based approaches and 
are restricted to a specific environment, they excel in robustness and can be applied even to a single image.
In this paper, we study PoseNet \cite{c3} and investigate modifications based on datasets' characteristics to improve the accuracy of the pose estimates. In particular, we emphasize the importance of field-of-view over image resolution; we
present a data augmentation scheme to reduce overfitting; we study the effect of  Long-Short-Term-Memory (LSTM) cells. Lastly, we combine these modifications and improve PoseNet's performance for monocular CNN based camera pose regression.
\end{abstract}
\vspace{-0,5cm}
\section{Introduction}
\vspace{-0,1cm}
The performance of many computer vision applications, such as autonomous vehicle navigation, augmented reality, and mobile robotics, heavily depends 
on good localization of the system with respect to its environment \cite{c18,c19,c20}. 
%
The recent success of CNNs in related tasks such as image classification and object detection \cite{c9,c4,c14} 
has led 
researchers to 
explore learning based solutions for place recognition~\cite{c15} and camera pose estimation \cite{c1}.
This seems promising given the ability of CNNs to learn high dimensional representations of the input data, automatically selecting
the optimal set of features to accurately regress the camera pose. 

One of the main obstacles for training a neural network for a supervised task such as camera pose estimation is the need for abundant labeled data. Fortunately, it has been demonstrated that 
transfer learning is effective in reducing the need for large labeled datasets \cite{c16}, \cite{c17}. In particular, representations learned by a CNN on a large image classification dataset can be fine-tuned to solve the camera pose estimation problem with much
smaller datasets.

The authors of PoseNet \cite{c1} leverage CNNs and transfer learning and propose a pure neural network based solution to 6-DoF camera pose estimation (i.e., 3D translation and 3D rotation) for a specific environment, addressing some limitations of traditional vSLAM algorithms \cite{c21}.
However, the accuracy obtained with this architecture is still significantly below what can be obtained with vSLAM methods, especially if the latter are trained on full sequences. 
In this study, we investigate how this gap can be reduced. In particular, we explore three possible causes, namely i) cropping of the input images, ii) overfitting to training data and iii) neglecting temporal information. Each time, we propose remedies to alleviate these shortcomings and evaluate their effectiveness according to each dataset's characteristics.
%
In particular, we show the importance of the input image's field-of-view in comparison to its resolution. 
Second, we reduce the extent to which overfitting affects the performance by introducing a specific scheme for Data Augmentation (DA). Third, we demonstrate the benefits of using Long-Short-Term-Memory cells (LSTMs) over Fully Connected layers (FC).
Finally, we incorporate all these techniques to improve PoseNet's performance for camera relocalization.

The remainder of this paper is organized as follows. Section~\ref{sec:related} describes related work. Next, in section~\ref{sec:background} we give more details on the PoseNet architecture, the loss functions we use and the datasets. Section~\ref{sec:method} contains the main contributions of our work. Section~\ref{sec:conclusions} concludes the paper.

\section{Related Works}
\label{sec:related}
Visual SLAM algorithms rely solely on the images coming from a camera, typically with a limited field of view, and do not use any other input such as GPS or inertial sensors. This remains an active area of research in the
computer vision community \cite{c22}, \cite{c26}. Although various solutions for vSLAM have been proposed for different environments and applications, many of them share a similar 
structure and therefore similar limitations \cite{c5}, \cite{c6}, \cite{c7}, \cite{c8}:
They often lose track 
due to motion blur, high speed rotations, partial occlusions and presence of dynamic objects in the scene. This makes them unsuitable for demanding applications such as localization of Unmanned Aerial Vehicles (UAVs). 
Besides, most visual SLAM algorithms rely on expensive pipelines that require a database of hand-crafted features, the camera's intrinsic parameters, a good initialization of the algorithm, 
selecting and storing key-frames and finding feature correspondences among images.
In addition, for monocular images, these approaches suffer from a  phenomenon known as scale drift where the scale of the objects in the environment cannot be accurately inferred, resulting in
inconsistent camera trajectories \cite{c28}.

To alleviate these problems an end-to-end trainable architecture for camera pose estimation called PoseNet is proposed \cite{c1,c2}. The authors of these works modified GoogleNet \cite{c11} and leveraged transfer learning from ImageNet \cite{c10} classification task to train a network for pose prediction using only monocular images. 
They further 
improved its performance in \cite{c3} by introducing more sophisticated loss functions for optimization.

\cite{c12} extends PoseNet using LSTM cells to better exploit the spatial information in each image. 
To this end, the last layer's features from GoogleNet are reshaped in 2D and the rows/columns of the corresponding matrix are fed to LSTM cells, one at each timestep. 
Finally, the cell's output for the last timestep is used to predict the 6-DoF pose.

LSTMs have also been used in \cite{c13} to exploit the temporal information between consecutive frames for a better localization accuracy. 
In this case, a bi-directional LSTM is fed with the features and pose information of the frames before and after the current frame to predict the current pose. 
One limitation of this approach is that using bi-directional LSTMs requires access to future frames at each timestep, which is is not possible in
real-time online applications. Furthermore, the performance is evaluated only for regressing the position and not the orientation.

Although PoseNet and its family of algorithms are not as accurate as vSLAM algorithms mentioned before, they work on monocular images and are shown to be more robust to motion blur and changes in  the lighting conditions. Furthermore, unlike traditional vSLAM solutions, they do no require access to camera parameters, good initialization and hand-crafted features. 

Posenet's successor architectures improve its performance by making its architecture more complex while neglecting the effect of the data on the final performance. In this paper, we try to improve the performance for PoseNet family of algorithms by targeting the information that can still be gained according to datasets' attributes without imposing a more complex CNN architecture.
Such modifications can be applied to the above mentioned works as well as those proposed in \cite{c29,c30,c31} to further improve their performance.


   \begin{figure}
      \centering
      \includegraphics[width=\linewidth]{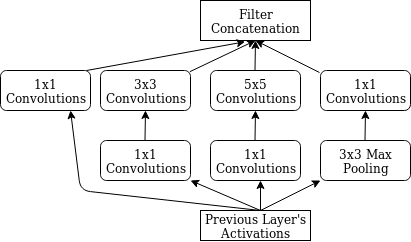}
      \caption{The inception module.}
      \label{inception}
   \end{figure}
   
      \begin{figure*}[thpb]
      \centering
      \includegraphics[scale=0.3]{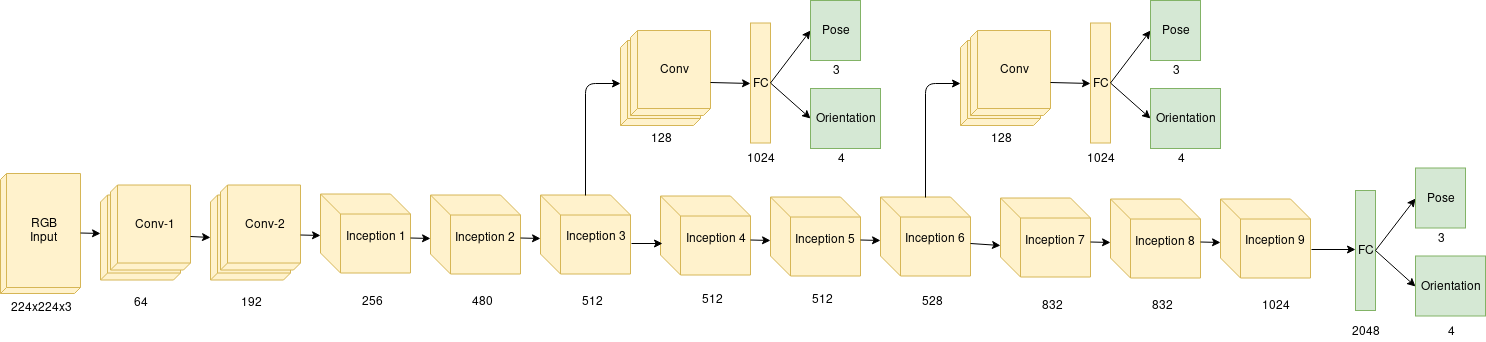}
      \caption{The Posenet architecture. Yellow modules are shared with GoogleNet while green modules are specific to Posenet.}  
      \label{posenet}
   \end{figure*}

\section{Background}
\label{sec:background}
In this section, we provide some background information on the architecture we used as the baseline for our experiments, largely building on~\cite{c3}. We modify this baseline in the next sections 
to improve the localization accuracy. 

\subsection{GoogleNet and Posenet}

Originally designed for object classification and detection, GoogleNet \cite{c11} 
is a 22 layer deep neural network based on a module known as Inception. An illustration of the inception module can be found in Figure \ref{inception}.
GoogleNet takes as input an image 
of $224\times224$ pixels
and propagates it through 9 inception modules stacked on top of each other using Rectified Linear Units (ReLu)
as the activation function.
Each layer in such a network learns a further abstraction of the input data. The highest level abstraction - which
resides on the last layer of the network - along with two intermediate abstractions are fed to fully connected and softmax layers to predict the objects' classes.

Posenet replaces these softmax classification layers with two parallel fully connected layers with 3 and 4 units respectively. These regress to pose (represented by $(x,y,z)$ coordinates) and orientation (represented as a quaternion).
Furthermore, 
a 2048 units fully connected layer is added on top of the last inception module. 
Figure \ref{posenet} illustrates PoseNet's architecture. The yellow blocks
represent the pretrained modules that PoseNet inherits from GoogleNet. The green blocks show PoseNet-specific modules that need to be trained from scratch. 

\subsection{Loss Function and Optimization}
The network described above, outputs a vector $x'$ and a quaternion $q'$ to represent the estimated position and orientation respectively. 
The parameters of the network are optimized for 
each image $I$ using the loss
function:
$$Loss(I)=||x-x'||_2+ \beta||q-q'||_2$$
where $x$ and $q$ represent the groundtruth position and orientation. Since quaternions are constrained to the unit manifold, the orientation error is typically much smaller than the position error. Therefore, a constant scale factor $\beta$ is used for balancing the loss terms.

The authors of PoseNet replace this constant scale factor with an adaptive one in \cite{c3}. The new loss function 
is formulated using homoscedastic uncertainty:
\begin{equation}
Loss(I)=||x-x'||_2\times e^{(\hat{-s_x})}+\hat{s_x}+||q-q'||_2\times e^{(\hat{-s_q})}+\hat{s_q}
\label{advanced}
\end{equation}
where $\hat{s} := \log\hat{\sigma}^2$ is a free scalar value trained by back propagation and $\hat{\sigma}^2$ denotes the homoscedastic uncertainty.
The uncertainty term $\hat{\sigma_q}^2$ is typically smaller than $\hat{\sigma_x}^2$, resulting in a larger weighting
factor for the orientation loss term and a well-balanced loss function.

For all experiments in this paper we optimize the above-mentioned adaptive loss function with the Adam optimizer \cite{c24}. The learning rate, $\beta_1$, $\beta_2$ and $\varepsilon$
for the Adam optimizer are set to 0.0001, 0.9, 0.999 and 1e-08 respectively.

\subsection{Datasets}
Following PoseNet and its successor architectures \cite{c3}\cite{c1}\cite{c2}\cite{c12}\cite{c13}, 
we report our results on the Cambridge Landmarks \cite{c1}
and 7-Scenes \cite{c25} 
datasets 
with median error values. It is worth mentioning that these two datasets differ both in their scale and data. 
Cambridge Landmarks consist of $1920\times1080$ images captured outdoors using a phone camera. The precise temporal information is lost as a result of the frame stream being sampled and many frames being removed. The labels for this dataset are produced using Structure from Motion.
The 7-Scenes dataset is recorded indoors using a Kinect RGB-D camera at a lower $640\times480$ resolution. All frames are kept and the labels are produced by a KinectFusion system \cite{c27}. 

The Cambridge Landmarks dataset covers larger areas in volume with relatively smaller number of frames compared to the sequences in the 7-Scenes dataset. Therefore, pose prediction in the
Cambridge Landmarks dataset is more closely related to place recognition while in the 7-Scenes dataset there is potential to exploit the temporal information for continuous localization.
In the following sections we sometimes refer to these datasets as outdoor and indoor datasets, respectively. Example frames for all sequences of both datasets are shown in Figure~\ref{trainingtest}.

\section{Method and Experiments}
\label{sec:method}

\subsection{Increased Field-of-View}
One drawback to transfer learning and using pretrained networks is that it imposes strong restrictions in terms of the network architecture. As an example, 
changing the input size is not an option.
Consequently, 
given that GoogleNet takes $224\times224$ RGB images as its input,
PoseNet resizes the smallest dimension of each image to 256 while keeping the original aspect ratio and crops a centered $224\times224$ window of the resulting image to train the network. In another training scheme, the network is trained on random crops of each image. In both of these cases localization accuracy is affected since the information outside cropping boundaries are lost.

Instead, we propose to use the entire field of view, by simply rescaling the input image to $224\times224$ pixels, even if that leads to a different aspect ratio.
We hypothesize that the loss of aspect ratio should not affect the performance of the network much,
since the new aspect ratio is consistent for all images in the dataset. Besides,  field-of-view is
of more importance compared to image resolution since the pooling layers in the network smooth the high frequency details of a higher-resolution image.

\begin{figure}
      \centering
    {\bf \small Original image} \hspace{0.8cm} {\bf \small PoseNet} \hspace{1.2cm} {\bf \small Ours} \hspace{1cm}
      \includegraphics[width=3.3cm,height=2.2cm]{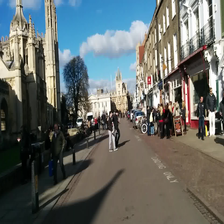}
      \includegraphics[width=2.2cm]{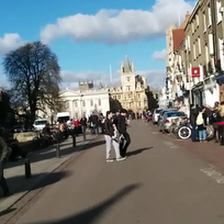}
      \includegraphics[width=2.2cm]{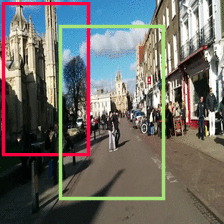}
      \\[1mm]
      \includegraphics[width=3.3cm, height=2.2cm]{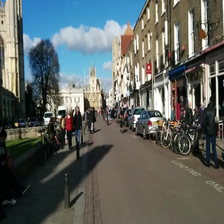}
      \includegraphics[width=2.2cm]{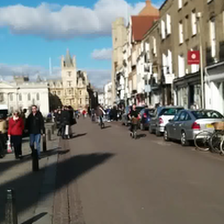}
      \includegraphics[width=2.2cm]{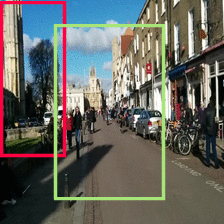}
      \caption{From left to right: Original input images, PoseNet's input and our input. PoseNet (middle/green rectangle) can lose important information for localization (highlighted in red).}  
      \label{fov}
   \end{figure}

Figure \ref{fov} shows an example of the difference in the field-of-view for the input to PoseNet and our network. 
The left column represents two images which their positions are 12 meters away from each other. The middle column shows the crop selected by PoseNet as the network's input while the right column illustrates our proposed alternative. 
As can be seen in the figure, PoseNet's field-of-view (middle column/the green rectangles in the right column) consists of dynamic objects and roughly similar distant landmarks which might not be very helpful for accurate localization. However, when considering the full field of view, as we suggest, closer and more useful landmarks for pose estimation, as highlighted by the red rectangles, come into view and can be exploited by the network.

Results in Table \ref{fovt} confirm that keeping the whole field of view in the image performs better than using a higher resolution crop with the original aspect ratio. This modification is specifically more in favor of the outdoor datasets where the area is larger and the landmarks close to the border of the image typically move faster than those in a centered crop. Therefore, such landmarks can play an important role for camera localization.
\begin{table}
\vspace{-0.3cm}
\begin{center}
\scriptsize
\begin{tabular}{|l||l||l|l|}
\hline
Dataset & Centered Crop & Whole Field of View & Improvement \\
\hline
King's College & 1.24m, 1.84$^{\circ}$ & {\bf 0.97m}, {\bf 1.27}$^{\circ}$ & 21.7\%, 30.9\%\\
Old Hospital & {\bf 3.07m}, 5.13$^{\circ}$ & 3.10m, {\bf 4.94} $^{\circ}$&-0.9\%, 3.7\%\\
Shop Facade & 1.23m,  5.74$^{\circ}$ & {\bf 0.93m}, {\bf 4.25}$^{\circ}$&24.3\%, 25.9\%\\
St Mary's Church & 2.21m, 5.92$^{\circ}$ & {\bf 1.66m}, {\bf 4.24}$^{\circ}$ &24.8\%, 28.3\%\\
Street & 21.67m, 32.8$^{\circ}$& {\bf 14.88m}, {\bf 24.35}$^{\circ}$ &31.3\%, 25.7\%\\
\hline
Average & 5.88m, 10.28$^{\circ}$&{\bf4.30m, 7.81}$^{\circ}$&20.2\%, 24.7\%\\
\hline
\hline
Chess & 0.18m, 5.92$^{\circ}$ & {\bf 0.16m}, {\bf 4.84}$^{\circ}$ &11.1\%, 18.2\%\\
Fire & 0.40m,  12.21$^{\circ}$& {\bf 0.35m}, {\bf 12.10}$^{\circ}$ &12.5\%, 0.9\%\\
Heads & 0.24m, 14.20$^{\circ}$& {\bf 0.20m}, {\bf 13.17}$^{\circ}$&16.6\%, 7.2  \%\\
Office & 0.30m, 7.59$^{\circ}$ & {\bf 0.25m}, {\bf 6.39}$^{\circ}$&16.6\%, 15.8\%\\
Pumpkin & 0.34m, 6.04$^{\circ}$ & {\bf 0.27m}, {\bf 5.53}$^{\circ}$ &20.5\%, 8.4\%\\
Red Kitchen & 0.36m, 7.32$^{\circ}$ & {\bf 0.30m}, {\bf 6.21} $^{\circ}$ &16.6\%, 15.1\%\\
Stairs & {\bf 0.35m}, 13.11$^{\circ}$  & 0.43m, {\bf 12.86}$^{\circ}$ &-22.8\%, 1.9\%\\
\hline
Average & 0.31m, 9.48$^{\circ}$&{\bf0.28m, 8.72}$^{\circ}$&10.1\%, 9.6\%\\
\hline
\end{tabular}
\end{center}
\vspace{-0.2cm}
\caption{Effect of increased field of view on localization accuracy 
}
\vspace{-0.4cm}
\label{fovt}
\end{table}
\subsection{Data Augmentation}

\begin{figure*}[thpb]
   \vspace{-0.8cm}
      \centering
      \includegraphics[width=0.88\linewidth]{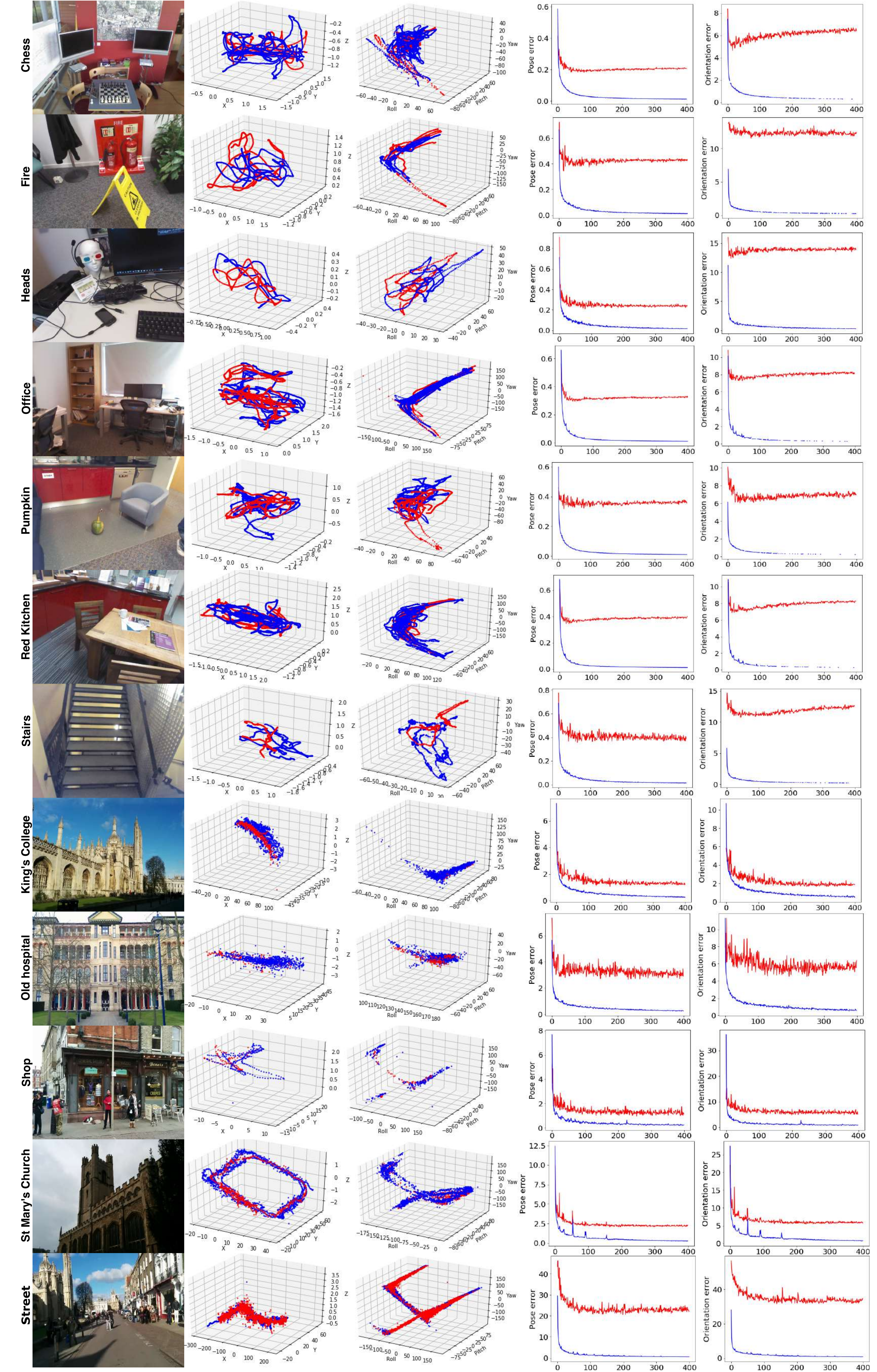}
      \caption{Left to right: an example image taken from each sequence, position trajectories for test (red) and training (blue) data, orientation trajectories, performance of the baseline model on test (red) and training data (blue) over different epochs, for pose and orientation.}
      \label{trainingtest}
   \end{figure*}

Figure \ref{trainingtest} demonstrates the performance of our baseline model-which uses Posenet's crop of the input image-
on training and test data for the different sequences.
As can be seen in this figure, the error on training data goes to zero after a small number of epochs while the error on test data
tends to be much higher. Besides, the performance on orientation prediction degrades over time for most of the indoor sequences which is a sign of overfitting to the training data.

\begin{table}
\begin{center}
\resizebox{\linewidth}{!}{%
\begin{tabular}{|l||l||l||l||l|}
\hline
Dataset &Baseline&Baseline-Augmented&Whole view-Augmented&Improvement (Column 2\&4)\\
\hline
King's College & {\bf 1.24m}, {\bf 1.84}$^{\circ}$ & 1.46m, 3.60$^{\circ}$&1.34m, 3.80$^{\circ}$&-8.0\%, -106.5\%\\
Old Hospital & 3.07m, {\bf 5.13}$^{\circ}$ & {\bf 2.56m}, 7.54$^{\circ}$&2.64m, 6.50$^{\circ}$&14.0\%, -26.7\%\\
Shop Facade & {\bf 1.23m}, {\bf 5.74}$^{\circ}$ & 1.31m, 7.14$^{\circ}$&{\bf 1.23m}, 7.33$^{\circ}$&0.0\%, -27.7\%\\
St-Mary's Church & 2.21m, 5.92$^{\circ}$ & 2.30m, 7.66$^{\circ}$&{\bf 1.80m},{\bf 5.85}$^{\circ}$&18.5\%, 1.1\%\\
Street & 21.67m, 32.8$^{\circ}$  & 17.63m, 34.86$^{\circ}$&\textcolor{green}{\bf 13.77m, 27.09}$^{\circ}$&36.4\%, 17.4\%\\
\hline
Average & 5.88m, 10.28$^{\circ}$&5.05m, 12.16$^{\circ}$&{\bf 4.15m}, {\bf 10.11}$^{\circ}$&12.1\%, -27.1\%\\
\hline
\hline
Chess & 0.18m, 5.92$^{\circ}$ & 0.20m, 5.08$^{\circ}$&{\bf 0.17m}, {\bf 4.65}$^{\circ}$&16.6\%, 21.4\%\\
Fire & 0.40m, 12.21$^{\circ}$& 0.37m, 9.86 $^{\circ}$&{\bf 0.34m}, {\bf 8.80}$^{\circ}$&15.0\%, 27.9\%\\
Heads & 0.24m, 14.20$^{\circ}$& 0.19m, 11.57$^{\circ}$&{\bf 0.16m}, {\bf 9.43}$^{\circ}$&33.3\%, 33.5\%\\
Office & 0.30m, 7.59$^{\circ}$ & 0.29m, 7.50$^{\circ}$&{\bf 0.27m}, {\bf 6.93}$^{\circ}$&10.0\%, 8.6\%\\
Pumpkin & 0.34m, 6.04$^{\circ}$ & 0.30m, 5.54$^{\circ}$&{\bf 0.24m},{\bf 4.86}$^{\circ}$&29.4\%, 19.5\%\\
Red Kitchen & 0.36m, 7.32$^{\circ}$ & 0.32m, 6.73$^{\circ}$&{\bf 0.29m}, {\bf 5.82}$^{\circ}$&19.4\%, 20.4\%\\
Stairs & {\bf 0.35m},13.11$^{\circ}$  & 0.42m, \bf{5.97}$^{\circ}$&{\bf0.38m}, 6.36$^{\circ}$&-8.5\%, 43.8\%\\
\hline
Average & 0.31m, 9.48$^{\circ}$&0.29m, 7.46$^{\circ}$&{\bf 0.26m}, {\bf 6.69}$^{\circ}$&16.4\%, 25.0\%\\
\hline
\end{tabular}}
\end{center}
\vspace{-0.2cm}
\caption{Data augmentation's effect on localization accuracy}
\vspace{-0.4cm}
\label{augment}
\end{table}

   \begin{figure}
      \centering
      \includegraphics[scale=0.35]{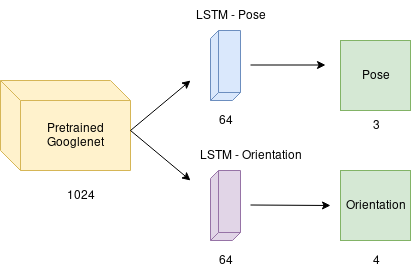}
      \caption{Proposed architecture for localization using LSTMs}  
      \label{lstm-arch}
   \end{figure}

As shown in Figure \ref{trainingtest} for many of sequences there is a limited overlap between the training
and test data for the orientation trajectories. This makes it hard for the network to extrapolate from the data it has seen during training given the limited number of training examples for each sequence.

Therefore, on each training epoch, we double the number of training examples by rotating each image with a random number in the range [-20,20] degrees. We manipulate the quaternion part of the label of the image to account for the new orientation. This way, we introduce new examples to the network with the same position 
but a different orientation. Therefore, this has the potential to improve the performance both for position and orientation. Table \ref{augment} shows the
performance of the proposed method.
A closer look at Figure \ref{trainingtest} and Table \ref{augment} suggests that generating new orientation labels is mostly useful for the indoor datasets where there is limited overlap 
between training and test orientation trajectories. However, for the outdoor sequences with enough overlap it can have a negative effect on the performance. 

It is worth mentioning that generating more training examples using a range other than [-20,20] or other data augmentation schemes such as horizontal, vertical flipping of the images can still further improve the results. However, the purpose of this section is to show overfitting as a problem for PoseNet and data augmentation as a remedy. Therefore further evaluation of the possible augmentation schemes are omitted.

   \begin{figure}
   \centering
      \includegraphics[width=\linewidth]{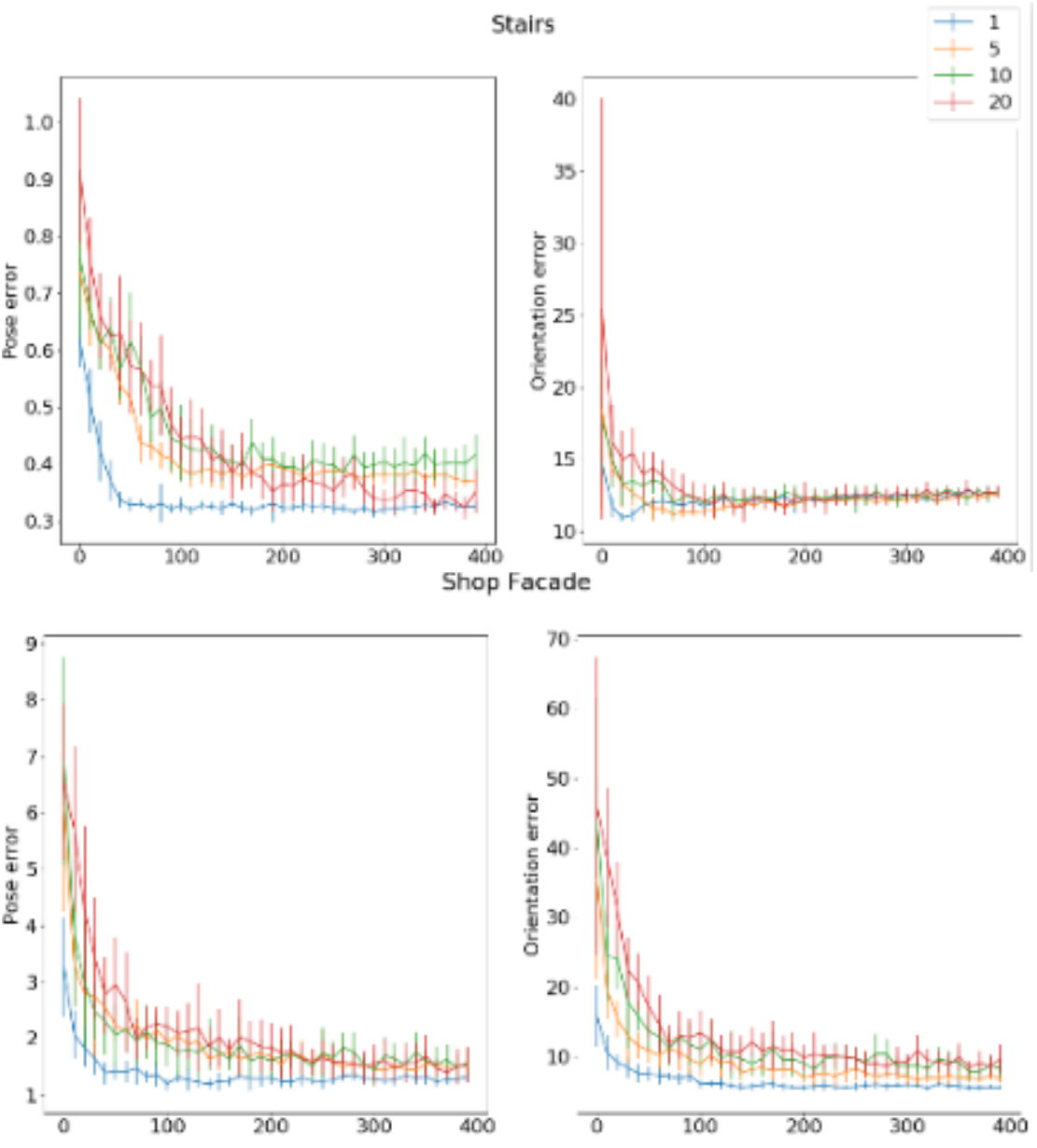}
      \caption{Results on Localization datasets using LSTM cells with different input sequence lengths}  
      \label{lstm}
   \end{figure}



 \begin{table}
\begin{center}
\resizebox{\linewidth}{!}{%
\begin{tabular}{|p{3cm}||p{2.5cm}||p{2.5cm}||p{2.5cm}||p{2.5cm}||p{2.5cm}|}
\hline
Dataset &Baseline&Lenght 1&Lenght 5&Lenght 10&Lenght 20\\
\hline
King's College & {\bf1.24m, 1.84}$^{\circ}$ & 1.32m, 2.08$^{\circ}$&1.48m, 2.66$^{\circ}$&1.79m, 2.77$^{\circ}$&2.05m, 3.28$^{\circ}$\\
Old Hospital & {\bf3.07m}, 5.13$^{\circ}$ & 3.09m, {\bf4.85}$^{\circ}$&3.38m, 6.35$^{\circ}$&3.79m, 5.79$^{\circ}$&4.05m, 7.36$^{\circ}$\\
Shop Façade & 1.23m, 5.74$^{\circ}$ & \bf{1.07m, 5.56}$^{\circ}$&1.25m, 6.86$^{\circ}$&1.41m, 7.39$^{\circ}$&1.28m, 9.68$^{\circ}$\\
St-Mary's Church & {\bf2.21m, 5.92}$^{\circ}$ & 2.32m, 8.00$^{\circ}$& 2.69m, 7.71$^{\circ}$&3.05m, 9.87$^{\circ}$&3.42m, 8.71$^{\circ}$\\
Street & {\bf21.67m, 32.8}$^{\circ}$  & 25.06m, 37.74$^{\circ}$&29.25m, 35.35$^{\circ}$&24.23m, 34.95$^{\circ}$&26.80m, 35.04$^{\circ}$\\
\hline
Average &5.88m, 10.28$^{\circ}$&6.57m, 11.64$^{\circ}$&7.58m,11.78$^{\circ}$&6.85m,12.15$^{\circ}$&7.52m,12.81$^{\circ}$\\
\hline
Chess & 0.18m, 5.92$^{\circ}$ & 0.19m, 6.20 $^{\circ}$&{\bf 0.16m, 5.88} $^{\circ}$& 0.17m, 7.07 $^{\circ}$&0.18m, 6.07$^{\circ}$\\
Fire & 0.40m, 12.21$^{\circ}$& 0.36m, 12.00 $^{\circ}$&{\bf 0.35m,10.95} $^{\circ}$& 0.39m, 12.38$^{\circ}$&0.38m, 12.21$^{\circ}$\\
Heads & 0.24m, 14.20$^{\circ}$& {\bf 0.18m}, 14.74$^{\circ}$&{\bf 0.18m},14.72$^{\circ}$&0.20m, 15.78 $^{\circ}$&0.22m, {\bf13.37}$^{\circ}$\\
Office & 0.30m, 7.59$^{\circ}$ & 0.32m, 8.35$^{\circ}$&{\bf 0.28m},7.46$^{\circ}$&0.29m, 8.39 $^{\circ}$&0.29m, {\bf7.04}$^{\circ}$\\
Pumpkin & 0.34m, 6.04$^{\circ}$ & 0.34m, 6.70$^{\circ}$&0.34m,6.39$^{\circ}$&0.34m, 7.29 $^{\circ}$&{\bf 0.32m, 5.88}$^{\circ}$\\
Red Kitchen & 0.36m, 7.32$^{\circ}$ & 0.35m,7.49$^{\circ}$&{\bf 0.32m},7.05$^{\circ}$&{\bf 0.35m},7.82 $^{\circ}$&0.35m, {\bf6.84}$^{\circ}$\\
Stairs & 0.35m,13.11$^{\circ}$  &0.34m, 11.85$^{\circ}$&0.36m,{\bf 10.83}$^{\circ}$&0.35m, 11.56$^{\circ}$&{\bf 0.33m}, 11.30$^{\circ}$\\
\hline
Average &0.31m, 9.48$^{\circ}$&0.29m, 9.61$^{\circ}$&{\bf 0.28m}, 9.04$^{\circ}$&0.29m, 10.04$^{\circ}$&0.29m, {\bf8.95}$^{\circ}$\\
\hline
\end{tabular}}
\end{center}
\vspace{-0.2cm}
\caption{Localization Accuracy Using LSTM Cells}
\vspace{-0.2cm}
\label{lstm-table}
\end{table}

 \begin{table}
\begin{center}
\resizebox{\linewidth}{!}{%
\begin{tabular}{|p{3cm}||p{2.5cm}||p{2.5cm}||p{2.5cm}||p{2.5cm}||p{2.5cm}|}
\hline
Dataset &Baseline&Lenght 1&Lenght 5&Lenght 10&Lenght 20\\
\hline
King's College & 1.24m, 1.84$^{\circ}$ &0.93m, 1.59$^{\circ}$&1.26m, 2.28$^{\circ}$&1.40m, 2.47$^{\circ}$&1.54m, 3.47$^{\circ}$\\
Old Hospital & 3.07m, 5.13$^{\circ}$ & 3.02m, 4.94$^{\circ}$&3.37m, 5.20$^{\circ}$&3.49m, 5.68$^{\circ}$&3.49m, 6.33$^{\circ}$\\
Shop Façade & 1.23m, 5.74$^{\circ}$ & \textcolor{green} {0.82m, 4.15}$^{\circ}$&1.31m, 7.87$^{\circ}$&1.60m, 6.95$^{\circ}$&1.44m, 11.84$^{\circ}$\\
St Mary's Church & 2.21m, 5.92$^{\circ}$ & 1.95m, 5.51$^{\circ}$& 2.07m, 5.56$^{\circ}$&2.28m, 7.03$^{\circ}$&2.53m, 8.10$^{\circ}$\\
Street & 21.67m, 32.8$^{\circ}$  & 16.48m, 30.00$^{\circ}$&18.31m, 33.61$^{\circ}$&20.93m, 34.42$^{\circ}$&26.80m, 35.04$^{\circ}$\\
\hline
Average &5.88m, 10.28$^{\circ}$&4.64m, 8.52$^{\circ}$&5.26m,10.20$^{\circ}$&5.94m,11.27$^{\circ}$&7.16m,12.95$^{\circ}$\\
\hline
Chess & 0.18m, 5.92$^{\circ}$ & 0.17m, 6.57 $^{\circ}$&0.15m, 5.84 $^{\circ}$&0.16m, 5.71 $^{\circ}$&0.15m, 6.02	$^{\circ}$\\
Fire & 0.40m, 12.21$^{\circ}$& 0.32m, 12.53 $^{\circ}$&0.32m, 12.65 $^{\circ}$&0.36m, 13.00$^{\circ}$&0.36m, 13.43$^{\circ}$\\
Heads & 0.24m, 14.20$^{\circ}$& 0.17m, 14.64$^{\circ}$&0.19m, 13.85$^{\circ}$&0.18m, 13.67 $^{\circ}$&0.20m, 14.53$^{\circ}$\\
Office & 0.30m, 7.59$^{\circ}$ & 0.28m, 8.65$^{\circ}$&0.27m, 8.96$^{\circ}$&0.24m, 8.30 $^{\circ}$&0.27m, 9.15$^{\circ}$\\
Pumpkin & 0.34m, 6.04$^{\circ}$ & 0.32m, 7.06$^{\circ}$&0.31m,6.89$^{\circ}$&0.36m, 8.08 $^{\circ}$&0.31m, 6.79$^{\circ}$\\
Red Kitchen & 0.36m, 7.32$^{\circ}$ & 0.29m, 7.07$^{\circ}$&0.28m, 6.91$^{\circ}$&0.26m, 5.94 $^{\circ}$&0.27m, 6.00$^{\circ}$\\
Stairs & 0.35m,13.11$^{\circ}$  & 0.30m, 11.46$^{\circ}$&0.36m, 10.19$^{\circ}$&0.33m, 11.69 $^{\circ}$&0.31m, 12.34$^{\circ}$\\
\hline
Average &0.31m, 9.48$^{\circ}$&0.26m, 9.71$^{\circ}$&0.26m, 9.32$^{\circ}$&0.27m, 9.48$^{\circ}$&0.26m, 9.75$^{\circ}$\\
\hline
\end{tabular}}
\end{center}
\vspace{-0.2cm}
\caption{Localization Accuracy Using LSTM Cells+Whole View (Green: Improves PoseNet based architectures, Refer to Table 5.)}
\vspace{-0.4cm}
\label{lstm-table-fov}
\end{table}

\subsection{LSTM experiments}
In order to exploit the temporal information between consecutive frames, we replaced the 2048 units fully connected layer in PoseNet with two parallel LSTM cells 
with 64 units for pose and orientation regression. This architecture is different from \cite{c12} where LSTMs are used for extracting spatial information 
and also different from \cite{c13} where stacked bi-directional LSTMs are used for pose-only regression. Figure~\ref{lstm-arch} illustrates the architecture we used for 
our experiments.

We fed sequences with 1, 5, 10 and 20 consecutive frames as input to our LSTM cells in separate experiments. 
Figure~\ref{lstm} illustrates the performance 
of our model on the test data, for one indoor and one outdoor localization scene. Experiments with different LSTM sequence lengths are 
shown in different colors in this figure. The results are averaged for every 10 epochs and the variance on every 10 epochs is shown with error bars. Tables \ref{lstm-table} and \ref{lstm-table-fov} show results on all localization datasets for different LSTM sequence lengths. The green colored result in the table improves the performance for PoseNet based architectures on the corresponding outdoor localization sequence (refer to Table 5 for results on those architectures).

While we expected to have high localization gains using longer sequence lengths, our results show that
LSTM cells with sequence length 1 seem to perform on average 
as good as longer sequence lengths. This behaviour is expected on the outdoor datasets
where the frames are downsampled and temporal information cannot be exploited.

For indoor datasets, this behaviour is justified by the fact that consecutive frames differ from each other with very small translations. CNNs trained on classification datasets such as ImageNet are designed to be translation-invariant and treat pose as a nuisance variable; therefore,
most of the temporal information is lost in such CNN.
The extracted features for consecutive frames
are always more than 98\% similar according to our experiments. This makes it hard for the LSTMs to differentiate consecutive frames.

Besides, with sequence length one, results using the LSTM architecture seem consistenly better than the CNN architecture, especially for the whole field-of-view setting.
Therefore, we conclude LSTM cells can improve localization performance independently from their sequence length due to their more complex architecture compared to fully connected layers.
It is worth mentioning that we reproduced the same results 
with different settings such as using stacked LSTMs, higher number of units for LSTMs and 
same LSTM to regress both pose and orientation.

\subsection{Putting it All Together}
Table 5 illustrates our localization performance compared to other works in the literature when we combine all the proposed modifications to the PoseNet baseline. The green colored numbers in the table suggest improvement over the performance of the PoseNet based works on the corresponding localization sequences.
 \begin{table}
\label{alltogether}
\begin{center}
\resizebox{\linewidth}{!}{%
\begin{tabular}{|p{3cm}||p{2cm}||p{2cm}||p{2cm}||p{2cm}||p{2.5cm}|}
\hline
Dataset &PoseNet&Spatial LSTMs \cite{c12}&Vidloc\cite{c13}&Posenet Adaptive Loss&This Paper\\
\hline
King's College & 1.66m, 4.86$^{\circ}$ & 0.99m, 3.65$^{\circ}$&-,-$^{\circ}$&0.99m, 1.06$^{\circ}$&1.45m,4.75$^{\circ}$\\
Old Hospital & 2.62m, 4.90$^{\circ}$ & 1.51m, 4.29$^{\circ}$&-, -$^{\circ}$&2.17m, 2.94$^{\circ}$&2.47m,5.64$^{\circ}$\\
Shop Facade & 1.41m, 7.18$^{\circ}$ & 1.18m, 7.44$^{\circ}$&-, -$^{\circ}$&1.05m, 3.97$^{\circ}$&1.13m, 7.35$^{\circ}$\\
St-Mary's Church & 2.45m, 7.96$^{\circ}$ & 1.52m, 6.68$^{\circ}$& -, -$^{\circ}$&1.49m, 3.43$^{\circ}$&2.10m,8.46$^{\circ}$\\
Street & -, -$^{\circ}$  & -, -$^{\circ}$&-, -$^{\circ}$&20.7m, 25.7$^{\circ}$&{\bf14.55m}, 36.04$^{\circ}$\\
\hline
Average &-, -$^{\circ}$&-, -$^{\circ}$&-,-$^{\circ}$&5.28, 7.42$^{\circ}$&{\bf4.34}, 12.44$^{\circ}$\\
\hline
Chess & 0.32m, 6.60$^{\circ}$ & 0.24m, 5.77 $^{\circ}$&0.18m, - $^{\circ}$&0.14m, 4.50 $^{\circ}$&0.17m, 5.34$^{\circ}$\\
Fire & 0.47m, 14.00$^{\circ}$& 0.34m, 11.9 $^{\circ}$&0.21m,- $^{\circ}$&0.27m, 11.8$^{\circ}$&0.30m, \textcolor{green}{10.36}$^{\circ}$\\
Heads & 0.30m, 12.2$^{\circ}$& 0.21m, 13.7$^{\circ}$&0.14m,-$^{\circ}$&0.18m, 12.1 $^{\circ}$&\textcolor{green}{0.15m, 11.73}$^{\circ}$\\
Office & 0.48m, 7.24$^{\circ}$ & 0.30m, 8.08$^{\circ}$&0.26m,-$^{\circ}$&0.20m, 5.77 $^{\circ}$&0.27m, 7.10$^{\circ}$\\
Pumpkin & 0.49m, 8.12$^{\circ}$ & 0.33m, 7.00$^{\circ}$&0.36m,-&0.25m , 4.82 $^{\circ}$&\textcolor{green}{0.23m}, 5.83$^{\circ}$\\	
Red Kitchen & 0.58m, 8.34$^{\circ}$ & 0.37m, 8.83$^{\circ}$&0.31m,-&0.24m, 5.52 $^{\circ}$&0.29m, 6.95$^{\circ}$\\
Stairs & 0.48m, 13.1$^{\circ}$  & 0.40m, 13.7$^{\circ}$&0.26m,-&0.37m, 10.6 $^{\circ}$&\textcolor{green}{0.30m, 8.30}$^{\circ}$\\
\hline
Average &0.44m, 9.94$^{\circ}$&0.31m, 9.85$^{\circ}$&0.24m, -$^{\circ}$&0.23m,7.87$^{\circ}$&0.24,7.94$^{\circ}$\\
\hline
\end{tabular}}
\end{center}
\vspace{-0.2cm}
\caption{Whole Field-of-View + Data Augmentation + LSTM}
\vspace{-0.4cm}
\end{table}
This table and the results discussed before suggest that for sequences where the frames are downsampled and there is enough overlap between training and test data,
the best performance is achieved by using the whole field-of-view as input. Otherwise, data augmentation can help  to reduce the margin between test and training data. Besides,
LSTM cells are generally helpful due to their relatively more complex architecture compared to fully connected layers, even when applied to sequences of lenght one, i.e. a single input image.
\section{Conclusion}
\label{sec:conclusions}
In this paper we study three different ways to improve camera localization accuracy of PoseNet targeting its training data's characteristics rather than network's architecture. Our experiments show that the field-of-view is typically more important than the input image's resolution. Furthermore, depending on the training labels' abundance one can benefit from data augmentation to cover more areas during training time to gain accuracy on test time. Besides, while LSTM cells do not seem to be able to exploit the temporal information due to the feature extraction network being invariant to small translations, they can perform better due to their more complex nature compared to fully connected layers.
\paragraph{Acknowledgment}
This work was supported by the FWO SBO project Omnidrone \footnote{https://www.omnidrone720.com/}.





\end{document}